\documentclass[conference]{IEEEtran}
\IEEEoverridecommandlockouts
\pdfoutput=1

\usepackage{cite}
\usepackage{amsmath,amssymb,amsfonts}
\usepackage{algorithmic}
\usepackage{graphicx}
\usepackage{textcomp}
\usepackage{xcolor}
\usepackage{svg}
\usepackage{hyperref}
\usepackage{tabularx}
\usepackage{tikz}

\def\BibTeX{{\rm B\kern-.05em{\sc i\kern-.025em b}\kern-.08em
    T\kern-.1667em\lower.7ex\hbox{E}\kern-.125emX}}

\newcommand\copyrighttext{%
  \footnotesize \textcopyright 2022 IEEE. Personal use of this material is permitted.
Permission from IEEE must be obtained for all other uses, in any current or future
media, including reprinting/republishing this material for advertising or
promotional purposes, creating new collective works, for resale or
redistribution to servers or lists, or reuse of any copyrighted
component of this work in other works.}
\newcommand\copyrightnotice{%
\begin{tikzpicture}[remember picture,overlay]
\node[anchor=south,yshift=10pt] at (current page.south) {\fbox{\parbox{\dimexpr\textwidth-\fboxsep-\fboxrule\relax}{\copyrighttext}}};
\end{tikzpicture}%
}    

\begin{document}

\title{SHOP: A Deep Learning Based Pipeline for Near Real-Time Detection of Small Handheld Objects Present in Blurry Video\\
}
\makeatletter
\newcommand{\linebreakand}{%
  \end{@IEEEauthorhalign}
  \hfill\mbox{}\par
  \mbox{}\hfill\begin{@IEEEauthorhalign}
}
\makeatother

\author{\IEEEauthorblockN{
Abhinav Ganguly\IEEEauthorrefmark{1},
Amar Gandhi\IEEEauthorrefmark{2},
Sylvia E\IEEEauthorrefmark{3},
Jeffrey Chang\IEEEauthorrefmark{4} and
Ian Hudson\IEEEauthorrefmark{5}}

\IEEEauthorblockA{\IEEEauthorrefmark{1}
Jacobs School of Engineering, 
University of California San Diego, San Diego, California, USA\\
aganguly@ucsd.edu}

\IEEEauthorblockA{\IEEEauthorrefmark{2}\
Department of Computer Science, 
California State Polytechnic University, Pomona, California, USA \\
acgandhi@cpp.edu}

\IEEEauthorblockA{\IEEEauthorrefmark{3}
College of Liberal Arts and Sciences, 
University of Illinois at Urbana Champaign, Champaign, Illinois, USA \\
sylviae2@illinois.edu}

\IEEEauthorblockA{\IEEEauthorrefmark{4}
Grainger School of Engineering, 
University of Illinois at Urbana Champaign, Champaign, Illinois, USA\\
jdchang3@illinois.edu}

\IEEEauthorblockA{\IEEEauthorrefmark{5}
Department of Mechanical Engineering, 
California State Polytechnic University, Pomona, California, USA\\
imhudson@cpp.edu}}


\maketitle


\copyrightnotice

\begin{abstract}
While prior works have investigated and developed computational models capable of object detection, models still struggle to reliably interpret images with motion blur and small objects. Moreover, none of these models are specifically designed for handheld object detection. In this work, we present SHOP (Small Handheld Object Pipeline), a pipeline that reliably and efficiently interprets blurry images containing handheld objects. The specific models used in each stage of the pipeline are flexible and can be changed based on performance requirements. First, images are deblurred and then run through a pose detection system where areas-of-interest are proposed around the hands of any people present. Next, object detection is performed on the images by a single-stage object detector. Finally, the proposed areas-of-interest are used to filter out low confidence detections. Testing on a handheld subset of Microsoft Common Objects in Context (MS COCO) demonstrates that this 3 stage process results in a 70 percent decrease in false positives while only reducing true positives by 17 percent in its strongest configuration. We also present a subset of MS COCO consisting solely of handheld objects that can be used to continue the development of handheld object detection methods. https://github.com/spider-sense/SHOP
\end{abstract}

\begin{IEEEkeywords}
machine learning, computer vision, object recognition, sharpening and deblurring
\end{IEEEkeywords}

\section{Introduction}
Though real-time object detection has become a widely studied field of computer vision, very little research has specifically addressed the detection of handheld objects. Many practical applications of handheld object detection involve visual security systems capable of weapon detection \cite{automated_firearms}. In scenarios like this, where time and accuracy are paramount, the effectiveness of a system is hampered by even small rates of false positives.

Research conducted on handheld object detection typically applies single-stage object detectors to images because they have a higher inference speed than two-stage detectors. However, single-stage object detectors have a lower mean average precision (mAP) than two-stage detectors and struggle with small handheld objects and blurry images due to a lack of object localization \cite{optimizing_trade_off}. In either case, a high confidence threshold would have to be used to obtain a low rate of false positives during handheld object detection, which would greatly diminish recall. 

The multi-stage pipeline we created improves the precision of real-time handheld object detection while maintaining a recall and runtime on par with most single-stage object detectors at a given threshold by filtering out low-confidence false positives. The pipeline first deblurs the image and proposes areas of interest around any detected hands. Then, object detection is performed on the image. Low confidence detections are tested for overlap with any areas of interest. Detections are eliminated if no overlap exists, or if the detection bounding box is too large to be a small object. 

When tested at various settings, our pipeline removed between 64-70\% of all false positives while removing between 13-33\% of all true positives. In its strongest performing configuration in \hyperref[table:results]{Table I}, the pipeline is able to remove 70\% of all false positives while only removing 17\% of all true positives. Our pipeline is also near real-time, running around 10 fps when tested on a Nvidia RTX 3060 GPU. 

The remainder of this paper is organized in the following way: \hyperref[sec:2]{Section II} overviews related works in object detection, pose estimation, and deblurring, highlighting current methods and some of their limitations. \hyperref[sec:3]{Section III} outlines the proposed pipeline and the motivations behind our design decisions. \hyperref[sec:4]{Section IV} goes over the creation of the datasets used for optimizing pipeline parameters, training, validation, and testing. Finally, \hyperref[sec:5]{Section V} reviews experiment results while \hyperref[sec:6]{Section VI} concludes with final thoughts and proposals for future work.

\section{Related Works}
\label{sec:2}
Here, we briefly summarize the related works to object detection and the different types of networks used. 

\subsection{Single and Two-Stage Object Detection}

In two-stage object detection, the locations of objects within images are estimated and mapped as regions of interest via a Region Proposal Network (RPN) \cite{feature_hierarchies} and are then used as input for a separate detection model. Single-stage object detectors like YOLO directly perform classification and regression on anchor boxes without the formation of regions of interest (ROI), which enables them to achieve high frames per second, but lower mAPs than two-stage detectors \cite{optimizing_trade_off}. 

\subsection{2D Pose Estimation}

2D human pose estimation processes visual data and returns a map with the location of body key points for any humans present, which connect to form a skeleton overlay over identified people. The two main approaches to 2D human pose estimation are top-down and bottom-up \cite{openpose}. The top-down approach identifies a single individual and creates their skeleton overlay one at a time. This is contrasted by bottom-up methods, where a body part map and a part relation 2D vector map identify and map all individuals present at once. Top-down approaches are more computationally expensive as their runtime scales linearly with the number of people present but are also more accurate compared to bottom-up approaches. 

Popular top-down approaches in pose estimation are SimDR \cite{sdr}, UDP-Pose-PSA \cite{psa}, OmniPose \cite{omnipose}, and HRNet-W48 \cite{MIPs}. Popular bottom-up approaches are Disentangled Keypoint Regression (DEKR) \cite{dekr}, SimplePose \cite{offsetguided}, and OpenPose (2017) \cite{openpose}. Notably, OpenPose is skilled at creating accurate maps even with high crowd occlusion. 

\subsection{Image Deblur via DeblurGANv2}

DeblurGANv2 \cite{deblurganv2} is a deep-learning based deblurring approach released by Kupyn et al. DeblurGANv2 uses a generic feature pyramid network for the network feature extractor backbone, allowing users a choice in backbones between Inception-ResNet-v2 \cite{inception_v4}, MobileNetV2 \cite{mobilenetv2}, and MobileNetDSC, which enables users to optimize for accuracy, speed, or a mix of both. MobileNetDSC and MobileNet's inference runtimes of $<$ 0.06 seconds makes real-time implementation a possibility. Inception-ResNet-v2 and DeblurGANv2 also achieve state-of-the-art results in the SSIM metric.

In particular, the original DeblurGAN \cite{deblurgan} paper found that implementing deblurring methods prior to performing YOLOv5 object detection on blurred images results in a larger F1 score, demonstrating heightened precision and recall on YOLOv5 object detection. This demonstrates that increasing the visual clarity of images can facilitate better object detection. Additionally, the DeblurGANv2 method was tested and determined to possess minimal deblurring artifacts, meaning that very few traces of deblurring are present in the deblurred image.

\section{Small Handheld Object Detection Pipeline}
\label{sec:3}
The pipeline consists of three main steps: deblur, object detection, and filtering. First, images are deblurred with DeblurGANv2. These deblurred images are processed by YOLOv5 to detect any present handheld objects. Lastly, these detections are collectively considered based on an area of interest generation system, where either SimDR or OpenPose is used to detect poses in the image and create areas of interest around the wrist.\footnote{OpenCV \cite{opencv_library} is used in intermediate processing steps of the pipeline.}

In the following subsections, each step is explained in detail. In the \hyperref[sec:addendum]{Addendum} section, we discuss how the pipeline's settings can be toggled to tailor it towards specific use cases. 

\begin{figure}[htbp]
\centerline{\includegraphics[]{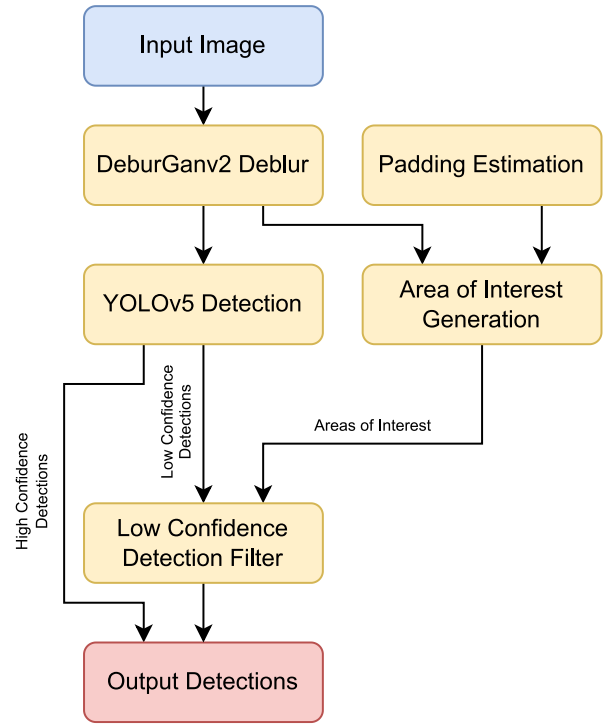}}
\caption{Pipeline process from input image to output detections}
\end{figure}
	
\subsection{Deblur}

Based on DeblurGAN's \cite{deblurgan} ability to facilitate increases in visual clarity, and subsequently higher F1 scores in object detection, we inferred that adding deblur as our first step would also enhance our pipeline’s pose-detection by allowing the pipeline to generate more relevant areas of interest during the filtering step. 

After reading a single image from the source, the pipeline uses an instance of the DeblurGANv2 \cite{deblurganv2} model with a MobileNetV2 feature extractor backbone to deblur the image. MobileNetV2 was used as our feature extractor backbone because it provides an order of magnitude decrease in runtime while maintaining an accuracy nearly on-par with the Inception-ResNet-v2 backbone \cite{deblurganv2}.

\begin{figure}[htbp]
\centerline{\includegraphics[trim={0 6cm 0 0},clip, scale=0.5]{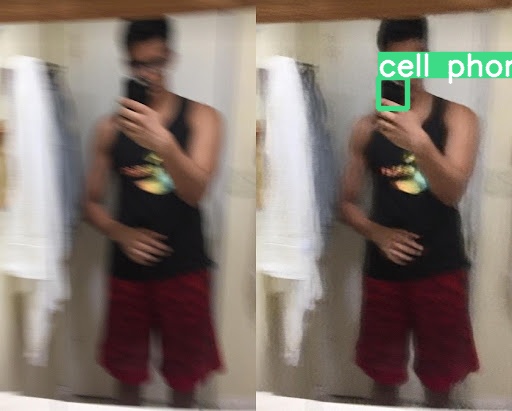}}
\caption{Phone was not detected in left image due to motion blur. Deblurring step allows phone to be detected on right image when running yolov5s.pt\cite{yolov5} at confidence threshold 0.6}
\end{figure}

\subsection{Object Detection}

After deblurring, the image is fed into the YOLOv5 object detection model to generate handheld detections. The training and specific classes used in this model are specified in \autoref{sec:4}. Detections that are above a user-defined upper confidence threshold are immediately accepted. If all detections in the image are above the upper confidence threshold, then the rest of the steps in this pipeline are skipped. Otherwise, the pipeline begins filtering out potential false positives from the group of detections by generating areas of interest. 

\subsection{Generating Areas of Interest}
\label{sec:aoi_creation}

To generate areas of interest, the pipeline first utilizes human pose estimation to obtain human key points, then uses the wrist points as center points for areas of interest, and finally creates a scaling factor to constrain the size of the areas of interest. 

\paragraph{Human Pose Estimation}
Prior to pose estimation, the pipeline performs human detection by running a YOLOv5m object detection model trained on the CrowdHuman dataset \cite{crowdhuman}. The number of humans detected is then used to determine whether to use SimDR or Openpose to detect the locations and human keypoints of people in an image.\footnote{We use Zhe Cao's TensorFlow OpenPose implementation \cite{tf_pose_est} and Sithu's SimDR implementation \cite{sithu_pose_est} in our code.} Whether SimDR or OpenPose is used is toggleable by the user: the pipeline will conditionally use SimDR, which will take the human bounding boxes as input, based on a user-defined cap on the number of humans in an image. If the cap is exceeded, the pipeline will switch to using OpenPose. 

The dual-use of these two pose estimation models in conjunction with a mutable transition limit allows users to tailor the pipeline towards speed or accuracy depending on the specific use case. SimDR demonstrates strong accuracy on COCO-test, but has a runtime that scales linearly based on the number of humans detected. In contrast, OpenPose demonstrates lesser accuracy than a top-down model like SimDR, but has a runtime invariant to the number of humans in an image. Consequently, our pipeline runs more quickly in images with more humans when compared to viable top-down alternatives such as UDP-Pose-PSA \cite{psa}, AlphaPose \cite{rmpe}, METU \cite{metu}, and SimDR \cite{sdr}.

The poses generated by this step are then used to generate areas of interest in the following steps.

\paragraph{Using Wrists as Center Points}

Among the key points generated by SimDR and/or OpenPose, the wrist was the most suitable candidate for a central keypoint. Wrists are the closest COCO keypoint to where handheld objects would be present, rarely deviate significantly in size and shape, and are less likely to be occluded by the presence of a handheld object. Wrists also solve many of the issues caused by directly detecting the hand, such as unreliable detection due to hand occlusion and variations in hand gestures \cite{handdet}. Generalizing the location of the hand based on an assumption that a vector is continued between the elbow and wrist keypoint would not work either as this generalization does not account for wrist flexibility in edge cases where the wrist bends perpendicularly to the elbow and wrist vector. 

In the event that the wrist is not detected or covered, the elbow is used as a backup center measurement. This provides the pipeline resiliency against situations where the handheld objects are held towards the camera (i.e. the wrist is hidden from view). In this situation, the distance between the wrist and elbow is insignificant and the occlusion of the wrist keypoint by the object implies that the handheld object is overlapping significantly with the human’s forearm and would likely be captured by generating an area of interest around the elbow keypoint.

\paragraph{Finding a Scaling Factor to Create Areas of Interest}

The pipeline uses the center wrist keypoint and its corresponding set of human keypoints to create areas of interest that filter out false positives. To create these measurements, we had to determine whether the detection is within 17.8 cm (7 in) of the wrist keypoint and the length of the detected handheld object.\footnote{Past research on hand lengths \cite{nasa} allowed us to determine that 18 cm is a sufficient distance from the wrist to the middle phalanx of each finger.} As the pixel distance representing 18 cm around a wrist varies based on the distance of a human from the camera, we needed a reliable scaling factor to approximate the number of pixels representing a single centimeter. This would be a portion of the human body that is generally consistent in width across people with different heights, genders, and physical orientations in an image.

The head fits these criteria. The head is invariant in width, roughly the same diameter in all orientations due to its cylindrical shape, and does not vary much in size.\footnote{Studies have shown that the difference between the 97th percentile of head circumferences of 200 cm tall men and the 3rd percentile of head circumferences of 150 cm tall men is about 10 centimeters, which translates to an approximately 3 cm (1.2 in) difference in head width \cite{head_centiles}.}
Thus, we used the average head diameter of 200 cm tall men of about 19.8 cm as our scaling factor; $\approx$19.8 cm are represented by the pixel width of the head.

In some cases, partial occlusion makes it hard to ascertain an accurate head width. In this situation, either the distance from the wrist to the elbow or from the elbow to the shoulder are used as a substitute scaling factor. These distances are best represented by the ulna and humerus bone length respectively.\footnote{The ulna bone is estimated to have a one standard deviation above average length of $\approx$26 cm (10.3 in) amongst men, and the humerus bone has a one standard deviation above average length of $\approx$32 cm (12.7 in) amongst men \cite{radius_anatomy}.} 

There were two issues we had to account for when using arm lengths as a scaling factor. First, arms can vary in significantly in size. To address this, we partially overestimated the arm length by assuming the subject was male and one standard deviation above average in arm lengths. The second issue was that perceived arm length can differ significantly based on the orientation of the arm relative to the camera. To address this, our pipeline first calculates the pixel widths of all detected arm key points, then only uses the largest pixel width arm as a scaling factor. This pixel width is least likely to grossly underestimate its arm’s actual length. If a forearm is chosen, the pipeline concludes that $\approx$26 cm are represented by forearm pixel width. If an upper arm is chosen, the pipeline concludes that $\approx$32 cm are represented by the upper arm pixel width.

If an accurate arm length cannot be ascertained, but both shoulder keypoints have been detected, the average shoulder width of $\approx$41 cm is used instead to represent the pixel distance between two shoulder keypoints.\footnote{The average shoulder width of men in the US is 41 cm \cite{shoulder_width}.}

The presence of large crowds in an image may cause OpenPose’s bipartite matching to misassign body parts to people. In particularly erroneous matchings, body parts used as scaling factors will not accurately represent a person's size, causing the pipeline to generate slightly smaller or excessively large scaling factors where a scaling factor is the pixel width divided by the number of centimeters it represents. While slightly smaller scaling factors do not greatly affect the pipeline’s accuracy, excessively large scaling factors can greatly increase the possibility of false positives. To address this, a ceiling was put on the potential length of the scaling factor when OpenPose is used by the pipeline.\footnote{This ceiling is not implemented when the top-down SimDR model is used, as keypoints are constrained within a human detection and so are far less likely to be misassigned between people.} If any arm length or head pixel width takes up over a fourth of the image’s max dimension, it is ignored. If the shoulder width pixel length takes up more than half of the image’s max dimension, it is ignored. If a scaling factor cannot be found, then the pipeline assumes that a fourth of the image’s max dimension represents 35.6 cm (14 in).

The scaling factor created by the above algorithm is then used to create areas of interest. They are represented as a list of bounding boxes surrounding the central wrist or elbow keypoint where each box represents a boxed region $\pm$18 cm (7 in) in height and width from the central point. 

\begin{figure}[htbp]
\centerline{\includegraphics[scale=0.15]{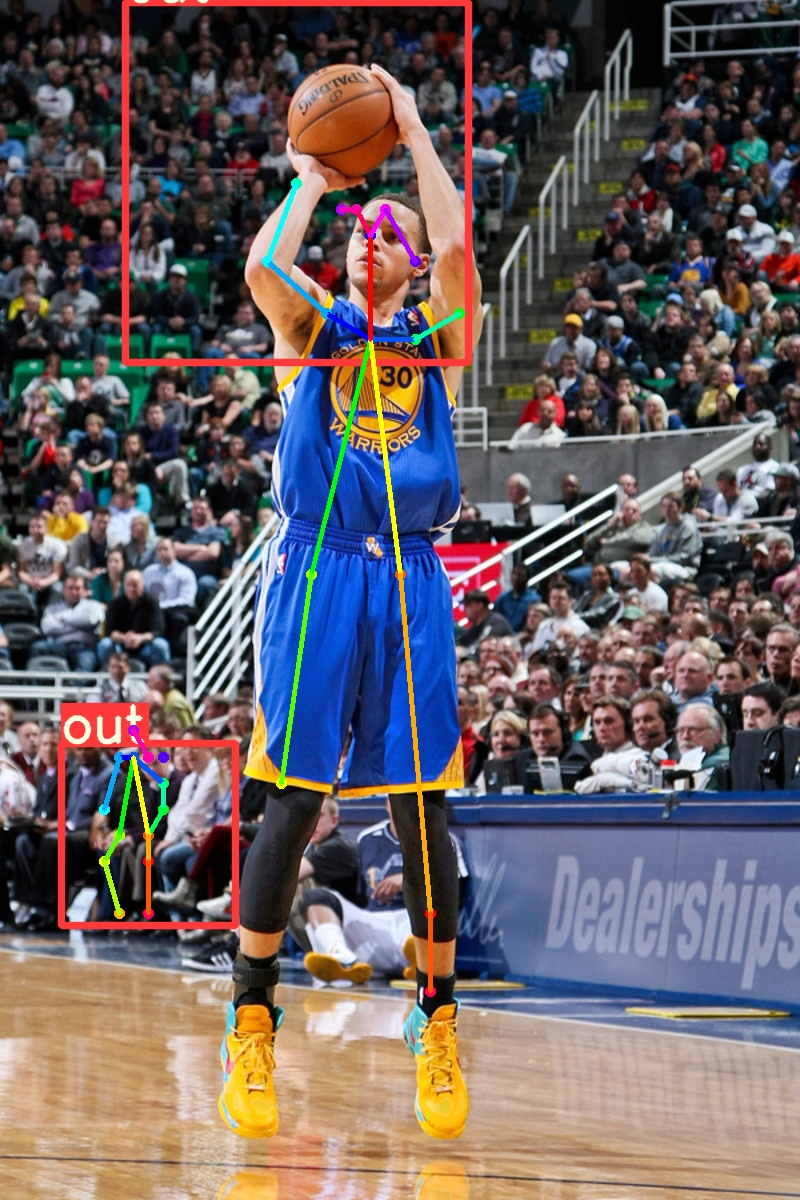}}
\caption{The pose detection corresponding to the person further away from the camera results in a smaller filter box (bottom left)}
\end{figure}

\subsection{False Positive Filtering}
\label{sec:fp_filtering}
After getting all detections and areas of interest, the pipeline iterates through the existing detections and only keeps those that meet one of the following conditions:

\begin{itemize}
\item Have a confidence value above the user-defined upper confidence threshold.
\item Have a user-defined percentage of its area overlap with one of the areas of interest boxes generated in \autoref{sec:aoi_creation}. Additionally, the maximum dimension of the detection cannot be greater than 2.5 times the width of the area of interest box.
\end{itemize}

The area of interest boxes are set as $\pm$18 cm (7 in), assuming that the handheld object will not be held more than 18 cm away from the wrist due to the wrist being adjacent to the hand itself. The 2.5 times maximum dimension cap puts an upper limit to the acceptable size of the detection. Based on the small handheld COCO classes specified in \autoref{sec:4}, we reasonably assumed that the vast majority of instances of these classes will be $<$53 cm (21 in).

Additionally, we added a user-defined upper confidence threshold in our pipeline. At higher confidences, most object detectors suffer from high rates of false negatives rather than high rates of false positives due to the stricter thresholds of accepting a detection \cite{pr}. As a result, applying the pipeline to high-confidence detections would not lead to a significant increase in precision and runs a higher risk of accidentally filtering out true positive detections. On a well-trained object detector, the majority of false positives are more likely to come from lower-confidence detections than higher-confidence detections so efforts were focused on filtering out the former.

\subsection{Addendum}
\label{sec:addendum}

The sections above highlight the general execution flow of our pipeline. Depending on user-defined settings for execution, the execution order may change. Users can choose to turn off deblurring if their system lacks the hardware to utilize DeblurGANv2 in concordance with pose-estimators and object detectors. In this scenario, the deblurring step is skipped. Additionally, the user may choose to remove the upper confidence threshold acceptance condition for detections highlighted in \autoref{sec:fp_filtering}. In this scenario, the execution order changes, and area of interest generation is performed before object detection. If no areas of interest can be generated, the pipeline skips the object detection stage and returns zero detections.

\begin{figure}[htbp]
\centerline{\includegraphics[scale=0.45]{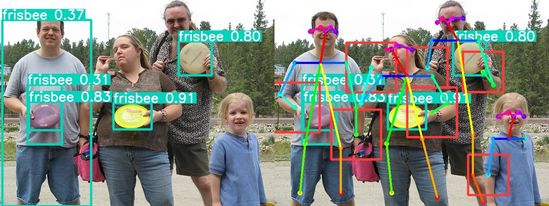}}
\caption{Running the pipeline removes one of the two false positives present in the left image. The person misclassified as a frisbee does not have 25\% of its area overlapping with a bounding box and is thus filtered out in the right image}
\end{figure}

\section{Dataset}
\label{sec:4}

To test the pipeline against a standard single-stage detection system, we needed a dataset that specifically focused on handheld objects. There are few datasets available for this task. The majority of relevant datasets, such as HOD \cite{hod} and SHORT \cite{short}, are focused on single object detection, where each object is fully centered in the image as opposed to “in the wild” detection of handheld objects. In order to further research into handheld object detection, we created our own dataset with train, validation, and test splits based on data from the COCO-train 2017 dataset \cite{mscoco}, an 80 class dataset of 118,287 images that is an industry-standard benchmark for object detection models.

The train split consists of 59,145 images (50\% of COCO-train) and the validation and test splits have 29,571 images each (25\% of COCO-train each). We also constructed a list of 26 classes (a subset of the 80 total classes in COCO) that represent small objects that can be handheld. 

\subsection{SHOP Dataset for future use}

To test end-to-end pipeline performance against a standard object detector, our evaluation program needed to be aware of whether an object is handheld or not. This way, the final evaluation didn’t penalize (or reward) the pipeline for missing (or correctly identifying) objects that were not being held. This meant the evaluation dataset needed to have both standard bounding box annotations and additional annotations for whether each bounding box was of a handheld or not handheld object. 

To construct the set, we started with the 29,571 images in the test split. Images without at least one instance of the person class were eliminated, as objects cannot be held without a person in the image to hold them. Images that did not have at least one instance of one of the 26 handheld classes discussed above were also eliminated. This left 6,501 images with 24,844 annotations for manual review. 

\begin{figure}[htbp]
\centerline{\includegraphics[scale=0.25]{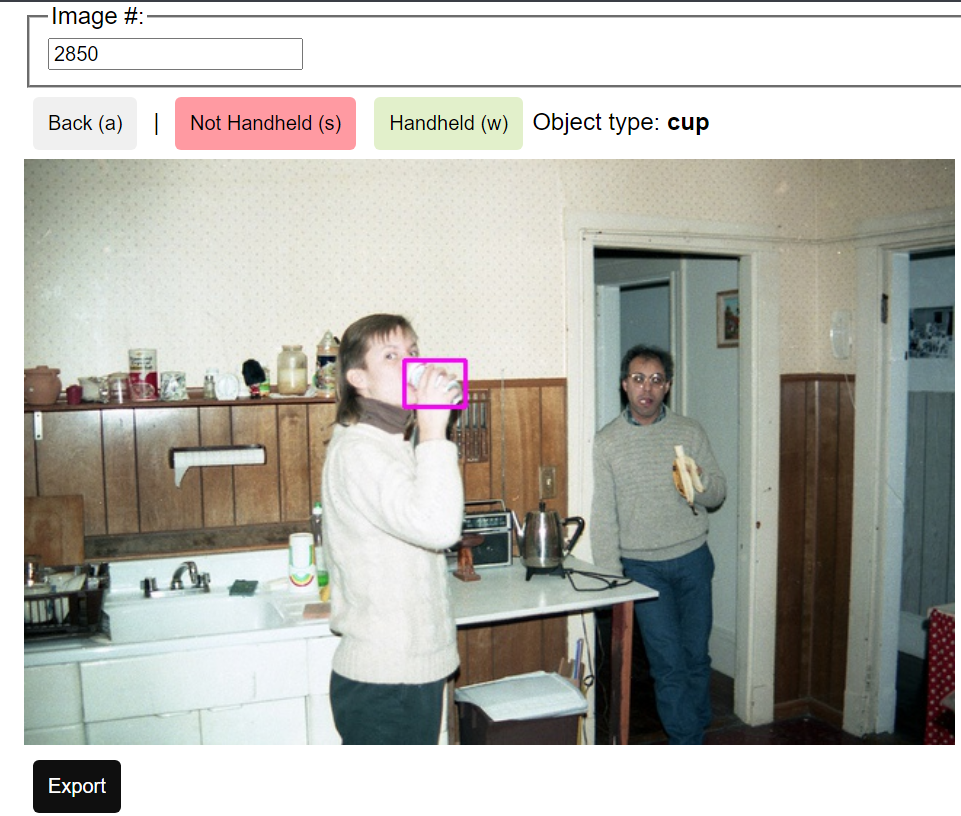}}
\caption{Image of personal annotation tool}
\end{figure}

Images were reviewed by unaffiliated workers to minimize the risk of labeling bias. To facilitate this, we created a simple website that showed annotations one at a time, with the option to mark each annotation as “handheld” or “not handheld,” along with an ID number for the current annotation they were looking at. Workers were assigned a range of annotation IDs to work on in blocks of 3,800. After labeling, the dataset had 3,716 images that contained at least one handheld object with 5,457 total instances of handheld objects. The final dataset is available at the following link: https://github.com/spider-sense/SHOP/releases. Due to time constraints, we were unable to use this dataset in the testing of our pipeline. We plan on utilizing the SHOP dataset in future research.

\subsection{Final Testing Set}
\label{sec:final_test}

Our pipeline tests were conducted against a handheld subset of COCO-validation, a 5000 image dataset. Similar to the COCO-train based testing set above, all non handheld objects in this set were eliminated. This left 1100 images with 1056 annotations of handheld objects.

\begin{table*}[t!]
\caption{
Results
\label{table:results}
}
\centering
\begin{tabular}{|c|c|c|c|c|c|}
\hline
\textbf{YOLO Weights}&\textbf{Pose Person Threshold}&\textbf{Upper-Conf Threshold}&\textbf{mAP}&\textbf{TP Filtered (\%)}&\textbf{FP Filtered (\%)} \\
\hline
yolov5s.pt  & Only top-down  & 1.1  & 0.294  & 25\%  & 64\%   \\ \hline
yolov5m.pt  & Only top-down  & 1.1  & 0.369  & 24\%  & 64\%   \\ \hline
yolov5l.pt  & Only top-down  & 1.1  & 0.393  & 24\%  & 64\%   \\ \hline
yolov5s.pt  & Top-down $\leq$ 3  & 1.1  & 0.267  & 33\%  & 69\%   \\ \hline
yolov5m.pt  & Top-down $\leq$ 3  & 1.1  & 0.335  & 32\%  & 69\%  \\ \hline
yolov5l.pt  & Top-down $\leq$ 3  & 1.1  & 0.357  & 32\%  & 70\%   \\ \hline
yolov5s.pt  & Only top-down  & 0.7  & 0.327  & 20\%  & 64\% \\ \hline
yolov5m.pt  & Only top-down  & 0.7  & 0.423  & 15\%  & 64\%  \\ \hline
yolov5l.pt  & Only top-down  & 0.7   & 0.456  & 13\%  & 64\%  \\ \hline
yolov5s.pt  & Top-down $\leq$ 3  & 0.7  & 0.306  & 27\%  & 69\%   \\ \hline
yolov5m.pt & Top-down $\leq$ 3 & 0.7 & 0.404  & 21\% &
 69\%  \\ \hline
 yolov5l.pt & Top-down $\leq$ 3 & 0.7 & 0.437  & 17\% &
 70\%  \\ \hline
\end{tabular}
\end{table*}

\begin{table*}[t!]
\centering
\label{table:prcurves}
\caption{PR Curves}
\begin{tabular}{|c|c|c|}
\hline
\multicolumn{3}{|c|}{
Pose with No Upper Confidence Threshold
}\\
\hline

\includegraphics[height=40mm]{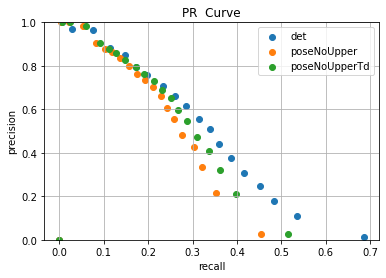} &
\includegraphics[height=40mm]{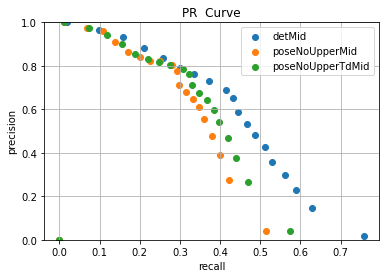} &
\includegraphics[height=40mm]{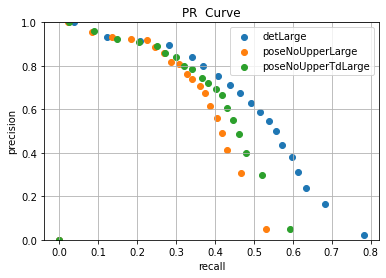} \\
YOLOv5s & YOLOv5m & YOLOv5l \\

\hline
\multicolumn{3}{|c|}{
Pose with Upper Confidence Threshold at 0.7
}\\
\hline

\includegraphics[height=40mm]{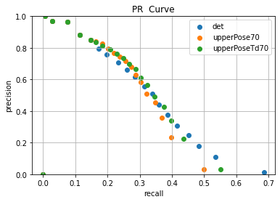} & 
\includegraphics[height=40mm]{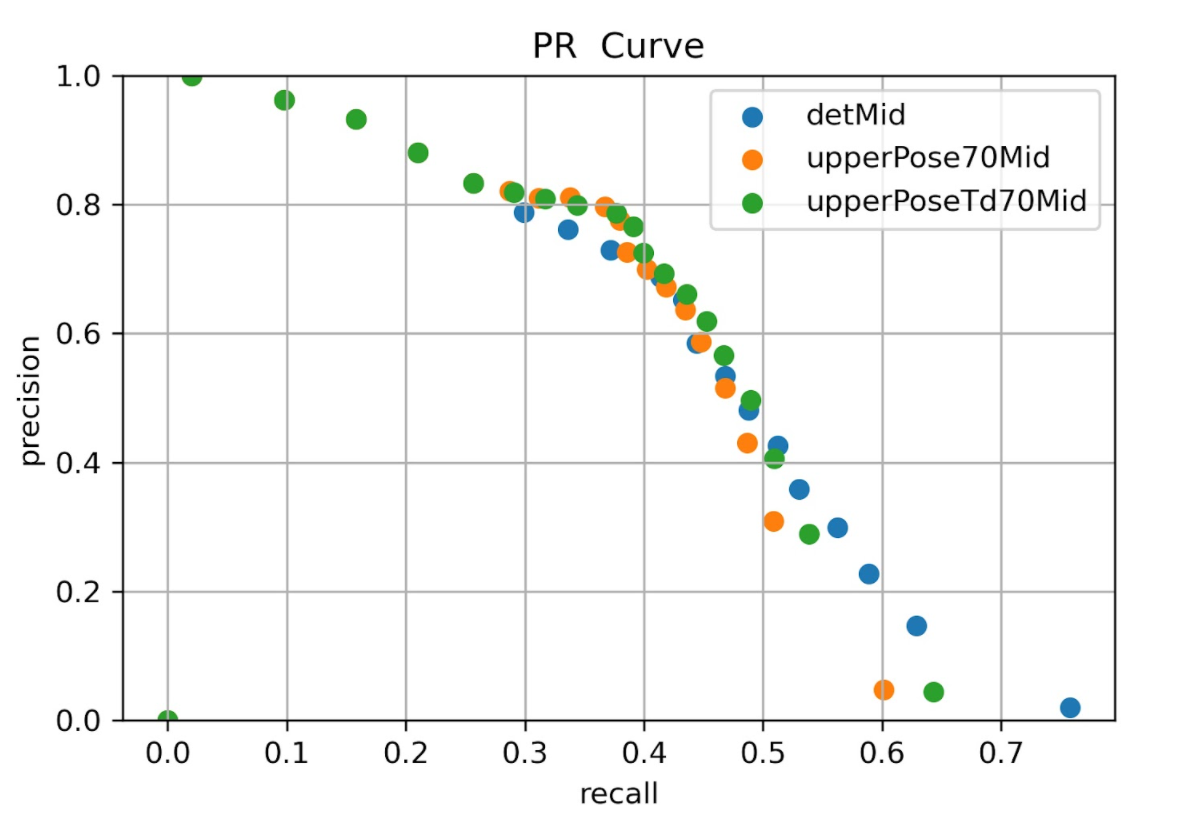} & 
\includegraphics[height=40mm]{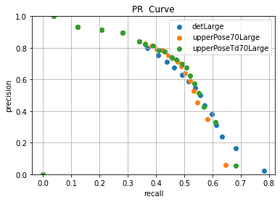} \\
YOLOv5s & YOLOv5m & YOLOv5l  \\
\hline
\end{tabular}

\small
\vspace{1ex}
\textit{
poseNoUpper = no upper confidence threshold, upperPose70 = upper confidence threshold at 70\%, \\ Td = only top-down, det = normal YOLOv5}

\end{table*}

\begin{table*}[t!]
\label{table:blurreddatasetresults}
\caption{Blurred Dataset Results}
\centering
\begin{tabular}{|c|c|c|c|c|c|}
\hline
\textbf{YOLO Weights}&\textbf{Pose Person Threshold}&\textbf{Upper-Conf Threshold}&\textbf{mAP}&\textbf{TP Filtered (\%)}&\textbf{FP Filtered (\%)} \\
\hline
yolov5s.pt  & Only top-down  & 0.7  & 0.320  & 20\%  & 64\% \\ \hline
yolov5m.pt  & Only top-down  & 0.7  & 0.420  & 15\%  & 64\%  \\ \hline
yolov5l.pt  & Only top-down  & 0.7   & 0.445  & 14\%  & 64\%  \\ \hline
yolov5s.pt  & Top-down $\leq$ 3  & 0.7  & 0.302  & 28\%  & 70\%   \\ \hline
yolov5m.pt & Top-down $\leq$ 3 & 0.7 & 0.400  & 21\% &
 69\%  \\ \hline
 yolov5l.pt & Top-down $\leq$ 3 & 0.7 & 0.424  & 19\% &
 70\%  \\ \hline

\end{tabular} 

\small
\vspace{1ex}
\textit{TP Filtered and FP Filtered are calculated by comparing the number of true positives and false \\positives filtered out when compared to YOLOv5 with a preprocessing deblur step}

\end{table*}

\begin{table*}[t!]
\centering
\label{table:prcurves2}
\caption{Blurred Dataset PR Curves}
\begin{tabular}{|c|c|c|}
\hline
\multicolumn{3}{|c|}{
Pose with Upper Confidence Threshold at 0.7
}\\
\hline

\includegraphics[height=40mm]{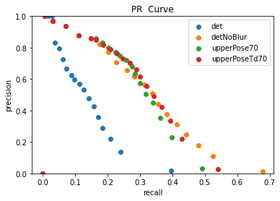} & 
\includegraphics[height=40mm]{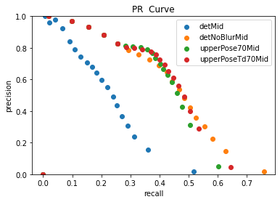} & 
\includegraphics[height=40mm]{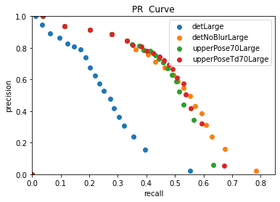} \\
YOLOv5s & YOLOv5m & YOLOv5l  \\
\hline
\end{tabular}

\small
\vspace{1ex}
\textit{Td = only top-down,
poseNoUpper = no upper confidence threshold, upperPose = upper confidence threshold at 70\%,
\\det = normal YOLOv5 run without deblurring, detNoBlur = normal YOLOv5 run with deblurring}

\end{table*}

\section{Results}
\label{sec:5}
While testing, we used the subset of COCO-validation specified in \autoref{sec:final_test}. All tests were run at a confidence of 0.001 with varying upper-confidence thresholds and pose thresholds of -1 and 3. This allowed half the tests to be run with only a top-down model (when pose threshold is -1); the other half were conditionally run using either a top-down or bottom-up model. Top-down models were utilized if 3 or fewer people were detected, while a bottom-up model was used when there were more than 3 people detected in an image. Images were set to require an overlap parameter of greater than 25\% with an area of interest generated box to be accepted. For this test, we did not utilize image deblurring as blur is typically not present in the COCO dataset. Both the baseline YOLOv5 and the pipeline were tested using YOLOv5s, YOLOv5m, and YOLOv51 models. We evaluated test runs by calculating the number of true positives, false positives, and false negatives outputted by the pipeline from confidences from 0.0 to 1.0 in intervals of 0.05. For each run, we plotted the PR Curve and calculated the mAP. After running tests, we also calculated the ratio of true positives our pipeline filtered out and the ratio of false positives our pipeline filtered out. The results of this experiment can be found in \hyperref[table:results]{Table I}.

For evaluation, we set the IOU threshold to 0.5. For each detection generated by the pipeline or by YOLOv5, the evaluation program tries to match it with a handheld object annotation. If it is matched, the detection is deemed a true positive. If the detection does not match a handheld object annotation, it is then compared to the original COCO annotations of the image, minus those that are present in the handheld objects list (essentially the set of non handheld objects in the image). Each handheld object annotation and COCO annotation can match against one detection, and each detection can match against one annotation. After completion of this process any remaining unmatched detections are deemed false positives, while any unmatched handheld annotations are deemed false negatives. Under this evaluation criteria, standard YOLOv5s detection received an mAP of 0.340, YOLOv5m detection received an mAP of 0.445, and YOLOv5l detection received an mAP of 0.479. 

Our pipeline displays a strong ability to filter out false-positive detections when tested on the validation set, filtering between 64-70\% of all false positives in all test cases while only filtering between 13-33\% of all true positives as a trade-off. However, this filtering ultimately leads to a rather significant dip in the mAP score of the pipeline when an upper confidence threshold is not inputted. This is likely because at higher confidences of detections, YOLOv5 suffers far more from a high rate of false negatives rather than a high rate of false positives. As a result, even though a larger percentage of false positives would be removed at these confidences, a larger number of true positives would be removed in total (as seen by the up to 33\% true positive filtering trade-off in \hyperref[table:results]{Table I}). Applying an upper confidence threshold remedies this issue, bringing the mAP to near the levels of the original pipeline. As can be seen by the Upper Pose PR curves in \hyperref[table:prcurves]{Table II}, our pipeline provides a higher mAP solution than YOLOv5 in an interval following the upper confidence threshold of 0.7. Our pipeline can offer higher precision and higher recall alternatives to YOLOv5 within this interval of confidence. 

To determine the effectiveness of a deblur step in filtering out false positives in blurry images, we conducted a second set of tests on an augmented version of this dataset. This blurry dataset was created by applying a randomized rotational and linear blur to images from the previous testing set.\footnote{Our code for creating the blur augmentation has been published on our linked repo.} We then conducted a similar test on this dataset with mostly the same pipeline settings and evaluation criteria as before with only a few minor changes. Firstly, we set the pipeline to deblur the image prior to finding areas of interest. Additionally, we ran a test where we ignored the pipeline and simply deblurred the image prior to running detections on it. The results of this test are shown in \hyperref[table:blurreddatasetresults]{Table III} and \hyperref[table:prcurves2]{Table IV}.

Our testing demonstrated that adding a deblur step to a detector in a blurry use case, even without area of interest generation, results in a significant boost in mAP. The mAPs for YOLOv5s, YOLOv5m, and YOLOv5l went from 0.147, 0.240, and 0.262 to 0.336, 0.441, and 0.476 respectively. Adding area of interest generation to the pipeline provides a similarly strong mAP bound around an upper confidence threshold of 0.7 as compared to a detector with just a deblur step. These results can be viewed in \hyperref[table:blurreddatasetresults]{Table III} and \hyperref[table:prcurves2]{Table IV}.

Overall, we believe that our pipeline demonstrates the legitimacy behind using deblurring and pose-based areas of interest generation to filter out location-based false positives particularly in blurry images. Using the dataset we created, we hope other researchers can build upon this work to provide a stronger filter that can provide greater boosts in mAP to handheld object detectors.

\section{Conclusion}
\label{sec:6}
The progression of handheld object detection research will greatly affect practical applications of object detection software. Throughout this paper, we presented a multi-stage pipeline that efficiently detects handheld objects at high precision while retaining a recall similar to that of single-stage object detectors. We demonstrated through validation testing that the localization of handheld areas of interest has the potential to filter out large amounts of location-based false positives resulting in a larger mAP bound around certain confidence thresholds and that the addition of a deblur step makes a detector significantly better in blurry use cases. In addition, we created an open-source dataset of handheld objects based on a subset of MS COCO where the objects are in candid positions rather than the focus of an image linked here: https://github.com/spider-sense/SHOP/releases. In future models, we anticipate taking steps to find and implement more advanced bottom-up and top-down models such as OffsetGuided and UDP-Pose for faster runtimes and more relevant areas of interest generation. More precise models would improve the true positive and false positive filtering rates, which would allow us to increase the confidence threshold at which our pipeline would be viable. Additionally, we would like to further test handheld detection models by training them on the SHOP dataset. We will also continue to look for methods to better localize locations of true positives and to filter out unwanted false positives. We anticipate that we will continue to use more modern neural networks and pipeline algorithms to reduce our computational cost and potentially make this pipeline real-time. These steps, in tandem, would likely increase the viability of our pipeline in real-life use cases.

\section*{Acknowledgment}
We would like to thank Dr. Ujjal Kumar Bhowmik, a professor with the Grainger College of Engineering at UIUC, for supporting us and giving us guidance through the process of writing this paper.

\bibliographystyle{IEEEtran} 
\bibliography{shop.bbl}

\end{document}